\documentclass{svproc}

\usepackage{url}

\usepackage{graphicx}
\graphicspath{ {./images/} }
\usepackage{float}
\usepackage{xcolor}
\usepackage{soul}
\usepackage{subcaption}
\usepackage{comment}
\usepackage[labelformat=parens,labelsep=quad,skip=3pt]{caption}

\begin{document}
\mainmatter              
\title{Careful with That! Observation of Human Movements to Estimate Objects Properties}
%
\titlerunning{Observation of Human Movements to Estimate Objects Properties}  
%
\author{Linda Lastrico\textsuperscript{*}\inst{1,}\inst{2}, Alessandro Carf\`i\inst{1}, Alessia Vignolo\inst{3}, Alessandra Sciutti\inst{3}, Fulvio Mastrogiovanni\inst{1} \and Francesco Rea\inst{2} }
\authorrunning{Lastrico et al.} 
%
\tocauthor{}

\institute{Dipartimento di Informatica, Bioingegneria, Robotica e Ingegneria dei Sistemi (DIBRIS), Università degli Studi di Genova, Genova, Italy\\
\and
Robotics, Brain and Cognitive Science Department (RBCS), Italian Institute of Technology, Genova, Italy\\
\email{linda.lastrico@iit.it}
\and
Cognitive Architecture for Collaborative Technologies Unit (CONTACT), Italian Institute of Technology, Genova, Italy
} 

\maketitle

\begin{abstract}
Humans are very effective at interpreting subtle properties of the partner's movement and use this skill to promote smooth interactions. 
Therefore, robotic platforms that support human partners in daily activities should acquire similar abilities. 
In this work we focused on the features of human motor actions that communicate insights on the weight of an object and the carefulness required in its manipulation. Our final goal is to enable a robot to autonomously infer the degree of care required in object handling and to discriminate whether the item is light or heavy, just by observing a human manipulation. This preliminary study represents a promising step towards the implementation of those abilities on a robot observing the scene with its camera. Indeed, we succeeded in demonstrating that it is possible to reliably deduct if the human operator is careful when handling an object, through machine learning algorithms relying on the stream of visual acquisition from either a robot camera or from a motion capture system. On the other hand, we observed that the same approach is inadequate to discriminate between light and heavy objects. 

\end{abstract}
\section{Introduction and background} 
In the context of human-robot interaction, a great effort is directed towards the development of the robot ability to understand implicit signals and subtle cues that naturally characterize human movements. This comes to have critical importance in situations where robots are used in unconstrained environments, for instance in manufacturing, helping human operators to lift loads or assisting elderly. In typical human-human interaction a considerable amount of information is exchanged through non-verbal signals, such as the attitude of an action, its tempo, the direction of the gaze and the body posture. It has been proved that people are able to correctly estimate the weight of an object, simply observing another person lifting it \cite{sciutti:weight}. Recent research confirmed that  the same information could be transmitted by a humanoid robot controlling the vertical velocity of its lifting movements \cite{PalinkoSciutti}. Humans manage to easily manipulate objects they have never used before: at first, by inferring their properties such as the weight, the stiffness and the dimensions also from the observation of others manipulating them; at a later time, using tactile and force feedback to improve the estimation. Replicating this behaviour in robotic systems is challenging. However, preliminary results have been achieved in estimating objects physical properties, relying on inference-based vision approaches \cite{billard:weight}.\\
The interaction with humanoid robots is particularly critical: driven by their appearance, humans tend to attribute those robots human-like abilities and, if their expectations fall short, the interaction may fail \cite{sandini}. Humans strongly rely on implicit signals to cooperate; therefore, in this context, to obtain seamless human-robot collaboration, humanoid robots need to correctly interpret those implicit signals \cite{legibility}. Furthermore, if we consider a scenario where the robot acts as helper or partner in an unconstrained environment, it acquires great importance to endow it with the ability of correctly estimating the characteristics of the handled objects; as a consequence, the robot can plan a safe and efficient motion action. In this study, we give particular attention to how a robot could assess an object features just by seeing it transported by a human partner. Inferring those properties from the human kinematics during the manipulation of the objects, rather than from their external appearance, grants the ability of generalizing over previously unseen items.
\subsection{Rationale} \label{Rationale}
Suppose to transport a glass full to the brim with water: the effort required to safely manage it without spilling a drop resembles the challenging scenario of porting an electronic device that could be damaged. If we want a robot to perform the same action, the first step would be to give the robot the capability of recognizing the intrinsic difficulty of the task; if we consider a hand-over task between a human and a robot, the latter should be aware that it is about to receive an object that requires a certain degree of carefulness in the handling. Moreover, an assessment of the weight of the object would allow an efficient lift. These features could be estimated from the human motion and ideally should be available before the end of the observed action, to trace the human abilities and to allow the robot to prepare for the possible handover. Differently from the weight, the concept of carefulness is not trivial. Previous studies have dealt with delicate objects, but focused more on robotic manipulation: the difficulty in the addressed tasks was given from the stiffness or the deformability of the item; tactile sensors where used for estimating the necessary force to apply a proper grasp \cite{sanchezGrip,grip}. In our study we consider the carefulness necessary to move an item from a wider perspective. Indeed, not only the object stiffness but also its fragility, the content about to be spilled, or its sentimental value may lead a person to perform a particularly careful manipulation. In those real-life cases we would like the robot to successfully estimate the carefulness required just by observing the human kinematics.
As a proof of concept, we recorded some transportation movements involving glasses which differed for weight and carefulness levels and some kinematic features, derived from the human motion, were used to train classifier algorithms. These features were obtained for comparison both from a motion capture system and a robot camera.
We hypothesize that, from the knowledge of kinematic features descriptive of the human movement: \textit{\textbf{(H1)}} it is possible to infer if carefulness is required to manipulate an object and \textit{\textbf{(H2)}} it is possible to discriminate a lighter object from a heavier one.
To validate our hypothesis we have collected a dataset of human motions while performing a simple transporting task; then we have trained state-of-the-art classifiers to determine if it is possible to distinguish the carefulness associated with an object and its weight, exclusively observing the human motion.

\section{Experimental setup} \label{experimental_setup}
\begin{figure}[h]
\begin{subfigure}{.5\textwidth}
  \centering
  \includegraphics[width=.99\linewidth]{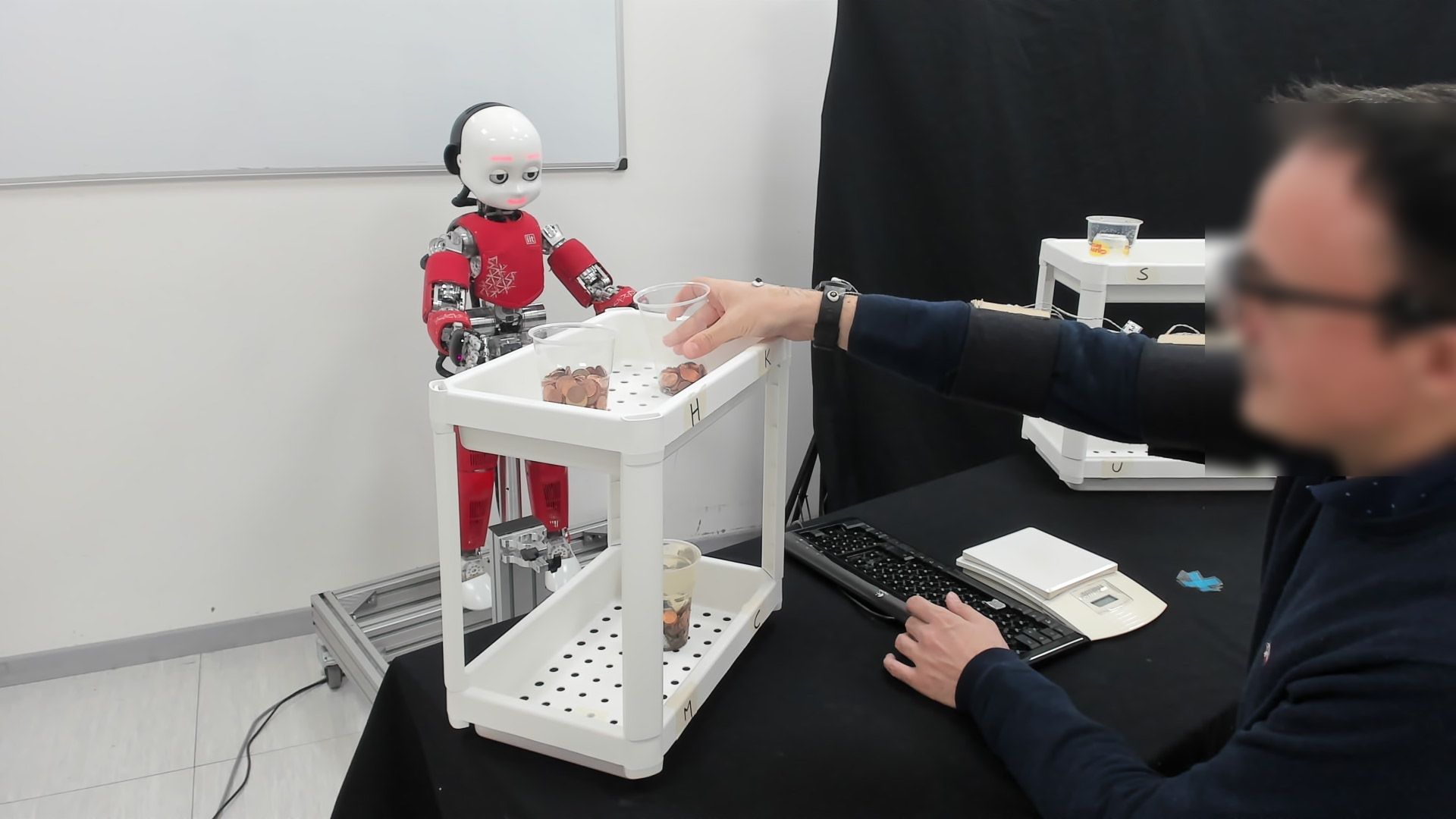}
  \caption{Lateral view}
  \label{fig:shelfreached}
\end{subfigure}%
\begin{subfigure}{.5\textwidth}
  \centering
  \includegraphics[width=.99\linewidth]{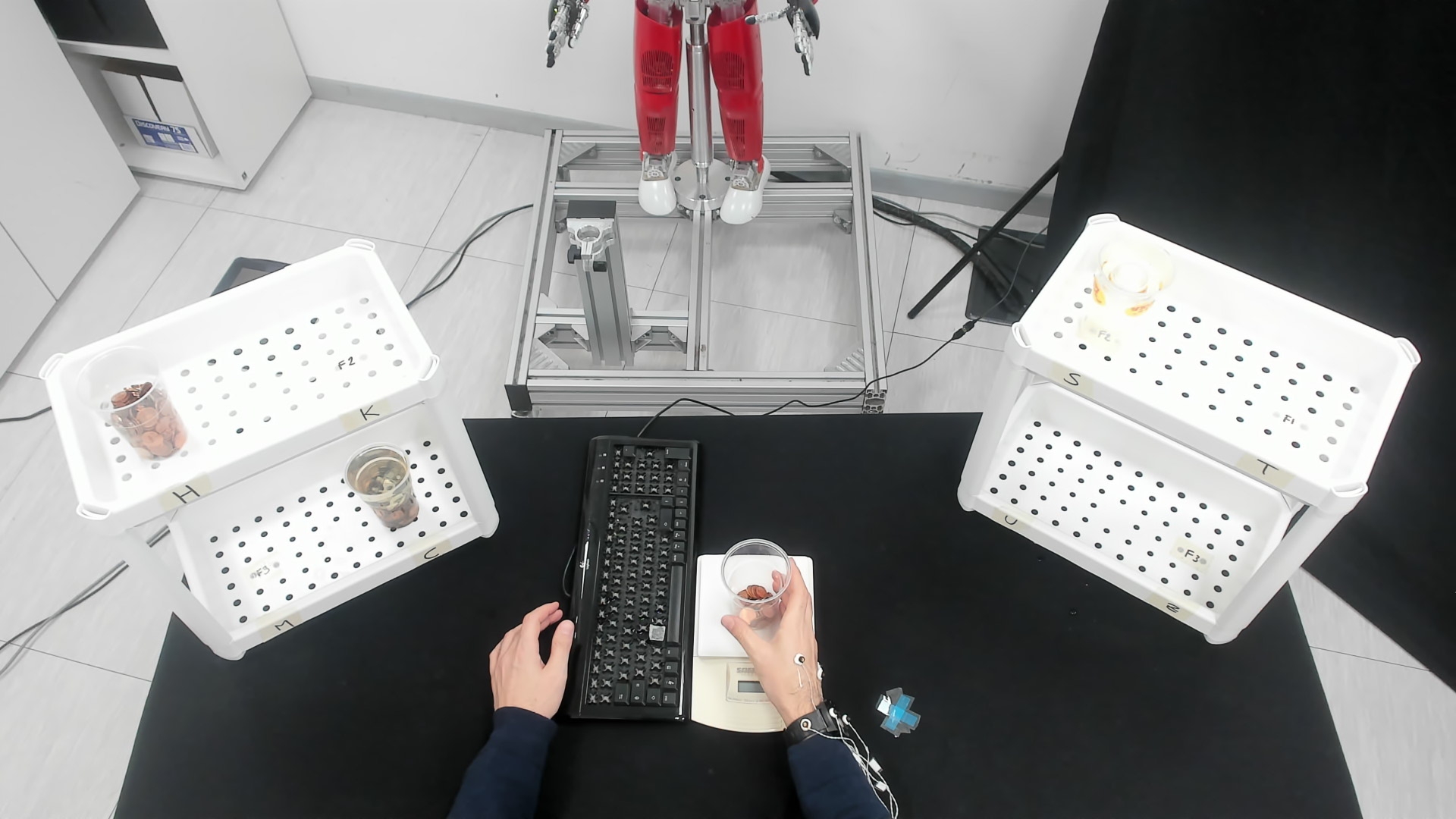}
  \caption{Top view}
  \label{fig:scalereached}
\end{subfigure}
\caption{Two views of the experimental setup with the volunteers in two grasp positions: on the shelf (\ref{fig:shelfreached}) and on the scale (\ref{fig:scalereached})}
\label{fig:setup}
\end{figure}

The experimental setup used to collect the data consisted of a table, a chair, two shelves (placed on different sides of the table) facing the volunteer, a scale, a keyboard with only one functioning key, and four plastic glasses (see Fig. \ref{fig:setup}). \\
%
\begin{table}[h]
  \begin{center}
    \caption{Glasses features and abbreviations}
    \vspace{0.1cm}
    \label{tab:table1}
    \begin{tabular}{c|c|c} 
      \textbf{Abbreviation } & \textbf{ Weight (gr) } & \textbf{ Carefulness level}\\
      \hline
      \hline
      W1C1 & 167 & low (no water)\\
      W2C1 & 667 & low (no water)\\
      W1C2 & 167 & high (full of water)\\
      W2C2 & 667 & high (full of water)\\
    \end{tabular}
  \end{center}
\end{table}
\\The four glasses were characterized by two levels of associated carefulness and two weights, as shown in Table \ref{tab:table1}. The high level of carefulness was achieved filling the glass to the brim with water, while for the low level no water was placed in the glass. The different weights, instead, were obtained by inserting in the glasses a variable number of coins and screws; for the object with high level of carefulness the weight of the water was taken into account. Each object was weighted to guarantee a difference of 500 gr between light an heavy glasses. Glasses were identical in shape and appearance, and their transparency was chosen so that participants could clearly see the content of the glass and appropriately plan their movements .
As displayed in Fig. \ref{fig:setup}, four positions were defined in each shelf, two on the top and two on the bottom level. These predefined positions were identified by a letter on a label. 


Participants seated at the table and had to perform a structured series of reaching, lifting and transportation movements of the four glasses. The experiment started with all the four glasses on the shelves, the volunteer with their arms resting on the table and their right hand in the resting pose, marked with a blue cross (see Fig. \ref{fig:scalereached}). During the experiment, the volunteers used their right hand to interact with the objects and their left to press the key of the keyboard. The experiment was structured as following: 
\begin{itemize}
    \item The volunteer pressed the key of the keyboard and a synthetic voice indicated the position on the shelf of the object to be transported. 
    The position was referred to using the corresponding letter.
    \item The volunteer performed a reaching action toward the specified position and grasped the glass (see Fig. \ref{fig:shelfreached}).
    \item The volunteer performed a transportation action moving the glass from the shelf to the scale.
    \item The volunteer released the glass and returned to the resting pose.
    \item The volunteer pressed a second time the key and the synthetic voice indicated a position on the shelf where the glass should be transported. 
    Of course, this time the selected position on the shelf was empty.
    \item The volunteer performed a reaching action towards the scale and grasped the glass (see Fig. \ref{fig:scalereached}).
    \item The volunteer performed a transportation action moving the glass from the scale to the final position on the shelf.
    \item The volunteer released the glass and returned to the resting pose.
\end{itemize}

The participants repeated this sequence 8 times to familiarize with the task, while the main experiment consisted of 32 repetitions. A table containing the shelf initial and final poses for each repetition was previously designed to guarantee a good coverage of all the possible combinations of shelf positions and glasses. Each volunteer performed exactly the same experiment.\\
The experiment was conducted thanks to 15 healthy right-handed subjects that voluntarily agreed to participate into the data collection (7 females, age: $28.6\pm 3.9$). All volunteers are members of our organization but none is directly involved in our research.
\subsection{Sensors}
\begin{figure}[ht]
\centering
\includegraphics[scale = 0.04, angle =-90]{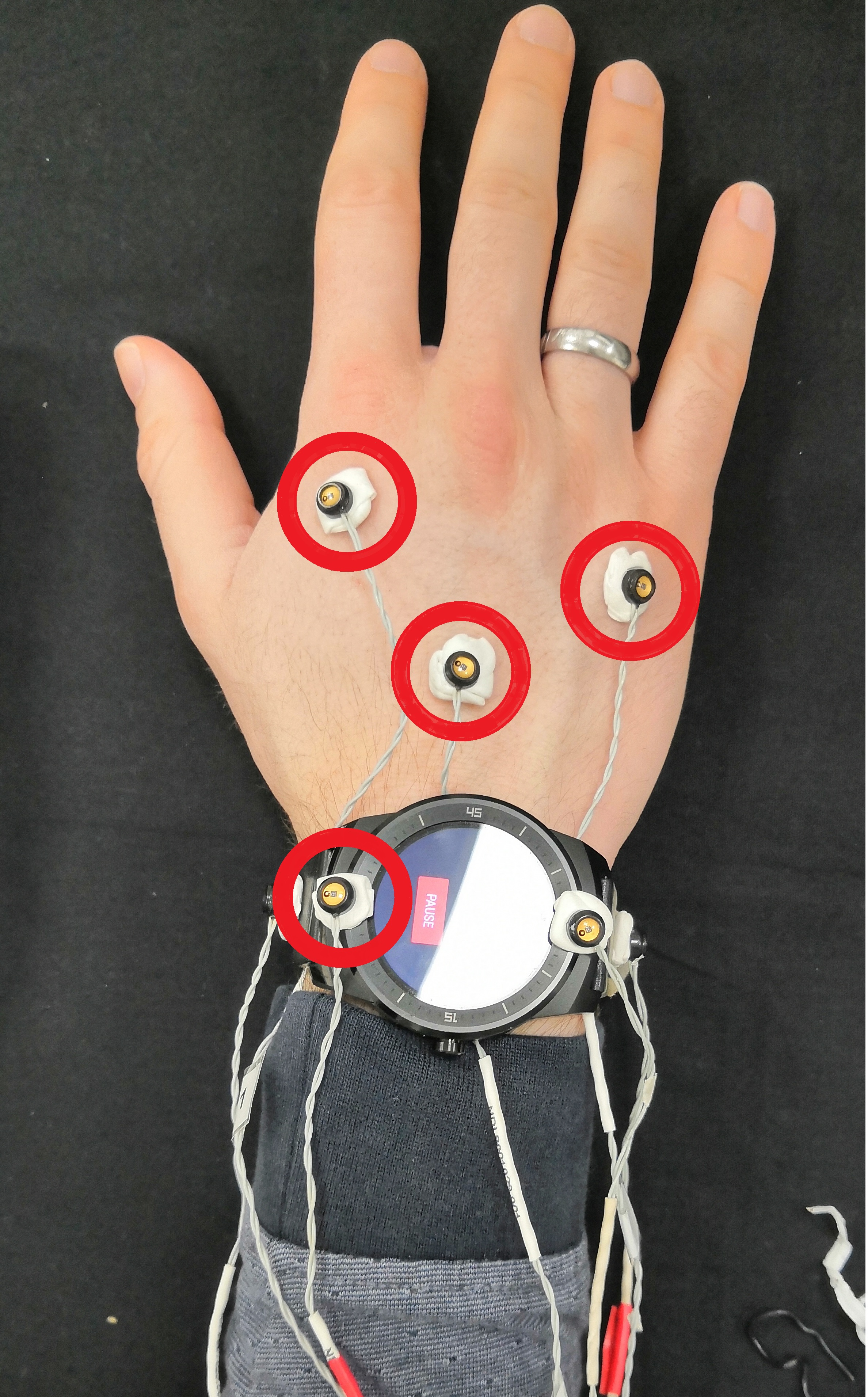}
\caption{Detail of the markers position on the right hand: those circled in red were interchangeably used to compute the features in each trial}
\label{fig:fig2}
\end{figure}

\begin{figure}[ht]
\centering
\includegraphics[width=0.8\textwidth]{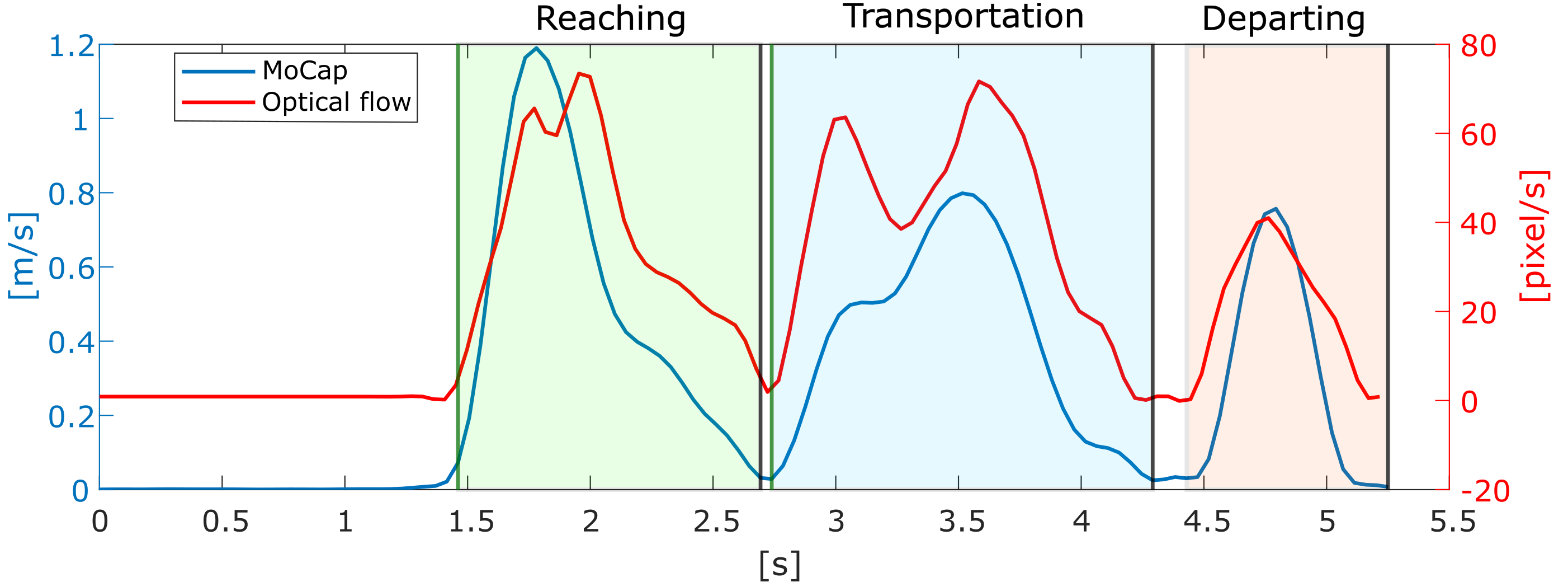}
\caption{Example of the velocity patterns from motion capture (in blue) and optical flow data (in red). The peaks characterizing the three phases of the trial (reaching, transportation and departing) are visible}
\label{fig:segmentation}
\end{figure}

The data used in this study was collected during the experiments previously described using a motion capture system from Optotrak, as ground truth, and one of the cameras of iCub. During the experiments other sensors have been used to collect data but their analysis is not in the scope of this paper.
The humanoid robot iCub was placed opposite to the table and recorded the scene through its left camera, with a frame rate of 22 Hz and a resolution of the image of 340 x 240 pixels. The robot was just a passive observer and no interaction with the participants took place during the experiment. The Optotrak Certus\textsuperscript{\textregistered}, NDI, motion capture (MoCap) system recorded the kinematic of the human motion through active infrared markers at a frequency of 100 Hz. The markers were placed on the right hand, wrist and arm. For the following analysis only a subset of the hand and wrist markers were considered (see Fig. \ref{fig:fig2}).
The data coming from the different sensors was synchronized through the middleware YARP \cite{yarp} that gave to each sample a YARP timestamp. 
By pressing the key on the keyboard at the end of every trial the data coming from the MoCap were automatically segmented in different log files and the actual timestamp saved in a separate file. Successively the timestamps associated with the key pressures have been used to segment the data recorded by the robot camera. 
\paragraph{Motion capture system data}
The data acquired by the motion capture system consisted in the tridimensional coordinates of each marker with respect to the reference coordinate frame of the Optotrak.
Occlusions limited the MoCap visibility for specific part of the human movement. In our experiment the main source of occlusion was given by the presence of the shelves, in particular for the lower right positions. To partially overcome this problem, after a preliminary analysis, we chose to consider for each trial the most visible marker among a subset of four as representative of the movement. Indeed, during the transportation movements the hand could be assimilated to a rigid body. The four considered markers were placed respectively on the metacarpophalangeal joints of the index and of the little finger, on the diaphysis of the third metacarpal and on the smartwatch in correspondence of the radial styloid (see the markers circled in red in Fig. \ref{fig:fig2} for reference). Two different interpolations, inpaintn \cite{inpaintn} and interp1 of MATLAB ver. R2019b, have been used to reconstruct the data that are missing because of the occlusions. The data was filtered with a second order low pass Butterworth filter with a cutoff frequency of 10 Hz.  Some trials have been excluded from the data set because of inconsistencies in the segmentation among the acquired sensors or because of errors of the subjects into pressing the key at the right moment, i.e. when their right hand was laying on the table in the resting position. Overall only 1.25\% of the total acquired trials have been removed.
Since our hypothesis is that it is possible to distinguish the features of the object that is being transported, it was necessary to isolate the transportation movement in every trial. To do so we took advantage of the experiment design. Indeed each trial presented three clearly identifiable phases: a reaching action, from the resting pose to the position occupied by the glass (either on the shelf or on the scale), a transportation movement and finally the departing (see Fig. \ref{fig:segmentation}). Our segmentation assumed that the start and end of the transportation phase is associated with a pick in the norm velocity of the hand. Therefore, the segmentation was performed by placing a threshold of 5\% on the second peak of the norm of the velocity, after filtering it with a fourth order filter with a cutoff frequency of 5 Hz. The resulting data were then down-sampled to obtain the same frame rate as the camera of the robot.

\begin{figure}
\centering
  \begin{subfigure}[b]{.325\textwidth}
    \includegraphics[width=1\linewidth]{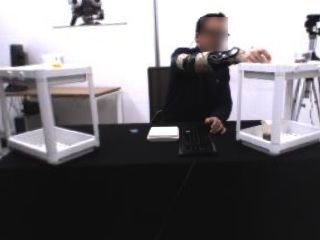}
    \caption{View from the iCub perspective}
    \label{fig:icubCam}
  \end{subfigure}
  \begin{subfigure}[b]{.325\textwidth}
    \includegraphics[width=1\linewidth]{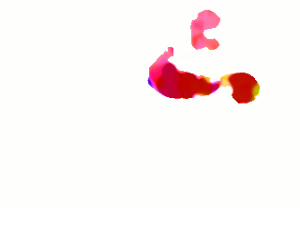}
    \caption{OF moving towards the right of the image}
    \label{fig:ofRight}
  \end{subfigure}
  \begin{subfigure}[b]{.325\textwidth}
    \includegraphics[width=1\linewidth]{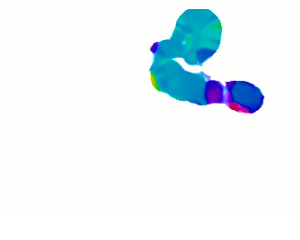}
    \caption{OF moving towards the left of the image}
    \label{fig:ofLeft}
  \end{subfigure}
  \caption{Example of iCub view of the scene and the extracted OF. The colors codify for the direction of the movement: red is for motion towards the right part of the image (\ref{fig:ofRight}), blue for motion towards the left (\ref{fig:ofLeft})}
  \label{fig:optiFlow}
\end{figure}

\paragraph{Camera data and optical flow extraction}
As motion descriptor, from the saved raw images of the robot camera (see Fig. \ref{fig:optiFlow} for an example) we chose to compute the Optical Flow (OF), following an approach already tested \cite{vignolo:OF,vignolo:OF2}. In this method, the optical flow is computed for every time instant using a dense approach \cite{farn:OF}, which estimates the apparent motion vector for each pixel of the image. The magnitude of the optical flow is thresholded to consider only those parts of the image where the change is significant. A temporal description of the motion happening in the derived region of interest is then computed averaging the optical flow components. On the velocity extracted, a second order low-pass Butterworth filter with cutoff frequency of 4 Hz was applied to remove the noise (see Fig. \ref{fig:segmentation}). 
\section{Data pre-processing} \label{feature_extraction}
The same set of motion representations was extracted during a pre-processing phase from both the motion capture data and the optical flow: the velocity $\mathbf{V}_{i}(t)$, the curvature $C_{i}(t)$, the radius of curvature $R_{i}(t)$ and the angular velocity $A_{i}(t)$ \cite{ActionObservation}. Their analytical expression is stated in Table \ref{tab:table2}. 
Such features can be computed for every time instant and by collecting them it is possible to progressively gather an increasing amount of information about the observed movement. This would then grant the robot the ability of discriminating online the characteristics of the object handled by the human partner. As shown in \cite{vignolo:OF,vignolo:OF2}, these data representations have been successfully used to discriminate online between biological and non-biological motion and to facilitate coordination in human-robot interaction \cite{Rea}. In addition, kinematics properties, such as velocity, have been shown to be relevant in human perception of object weight \cite{velWeight}. Extracting those  features during the pre-processing, instead of directly feeding the classification algorithms with raw data, allows to better compare the performance achieved with the two sources of data. Indeed, a precise control over the information used during the learning process is granted.

\begin{table}[h!]
\caption{Motion features computed from motion capture and optical flow data}
\centering
\label{tab:table2}
\renewcommand\arraystretch{2}
\begin{tabular}{l|l}
\textbf{Motion feature} & \textbf{ Analytical expression } \\
\hline
\hline
Tangential velocity & $\mathbf{V}_{i}(t) = (u_{i}(t),v_{i}(t),\Delta _{t})$            \\
\hline
Tangential velocity magnitude\hspace{0.2cm} & $V_{i}(t)=\sqrt{u_{i}(t)^{2}+v_{i}(t)^{2}+\Delta _{t}^{2}}$\\
\hline
Acceleration & $\mathbf{A}_{i}(t)=(u_{i}(t)-u_{i}(t-1),v_{i}(t)-v_{i}(t-1),0)$            \\
\hline
Curvature           & $C_{i}(t)=\frac{\left \|\mathbf{V}_{i}(t)\times \mathbf{A}_{i}(t) \right \|}{\left \| \mathbf{V}_i(t)) \right \|^{3}}  $ \\
\hline
Radius of curvature & $R_{i}(t)=\frac{1}{C_{i}(t)}     $              \\
\hline
Angular velocity    & $A_{i}(t)=\frac{V_{i}(t)}{R_{i}(t)}    $     \\
\hline
\end{tabular}
\end{table}
\subsection{Dataset} \label{dataset}
As we have detailed before some sequences had to be removed for inconsistencies in the segmentation. This lead to a slightly unbalanced data set, containing more examples for specific classes. Indeed, class W1C1 had 235 sequences, class W2C1 239, class W1C2 238 and class W2C2 had 236. Although cardinally the difference is minimum, to preserve the balance of the dataset we decided to fix the maximum number of sequences for each class to 235 and we have randomly selected the sequences for W2C1, W1C2 and W2C2. Notice that the four classes were characterized only by the weight and the carefulness level. Therefore other variables, such as the initial and final position of the glass and the direction of the movement, are not considered in the classification.\\
Due to the characteristics of the glasses, the duration of the transport movement varied consistently among the trials (i.e. the duration of the movement is consistently longer when the moved glass is full of water, belonging to the high carefulness class). To obtain sequences with the same number of samples for each trial, the segmented sequences were re-sampled, using the interp1 function of MATLAB. The number of samples was selected considering the class associated with the shorter duration of the transport phase, W1C1, and computing the median value among all its trials. The resulting value was 32. Therefore, our dataset was composed of two data structures: one derived from the MoCap data and the other one from the OF. Both structures had dimensions $940\, (trials) \times 32\, (frames) \times 4\, (features)$.\\
The re-sampling can be performed only knowing the start and end of the transportation phase. Since in an online scenario this information is not available, a further attempt was performed exploiting the ability of certain models to handle temporal sequences of different lengths. In this case, instead of re-sampling, a common zero-padding and masking technique were adopted. Therefore, the shorter temporal sequences were completed with zero values and those values were then ignored during the training, while the length of the longest transport movements was preserved. The shape of the data structures after the zero padding was: $940\, (trials) \times 132\, (frames) \times 4\, (features)$.

\section{Classifiers} \label{classifier}
As introduced in Sect. \ref{Rationale}, the goal of the classification is to discriminate between the two possible features of the transported glasses: \textit{\textbf{(H1)}} the carefulness level associated with the object and \textit{\textbf{(H2)}} the weight. Therefore, we decided to approach the problem using two binary classifiers, one for each feature, implemented in Python using Keras libraries \cite{keras}. As mentioned in Sect. \ref{dataset} two models were tested: the first one relied on re-sampled features, while the second one used the original data with variable lengths.
\subsection{Convolutional, Long-Short-Term-Memory and Deep Neural Network}\label{CNN}
Previous literature suggests that the combined use of Convolutional Neural Network (CNN), Long-Short Term Memory (LSTM) and Deep Neural Networks (DNN) is a good solution for classifying time dependent data, such as speech or motion kinematics \cite{mymodel,lstm}. Therefore, our first model was inspired by \cite{mymodel} and consisted of two time distributed 1-D convolutional layers (that took as input 4 subsequences of 8 frames each), a max pooling and flatten layers, a 100 neurons LSTM, a 100 neurons Dense layer and a 2 neurons output layer with a sigmoidal activation function. A Leave-One-Out approach was adopted, to test the ability of the model to generalize over different participants. Thus, for each one of the 15 folds, the data from 14 subjects were used as training set and the data of the fifteenth participant as test set. The 20\% of the data for each training set was kept for validation, and early stopping was implemented according to the validation loss function (with a patience parameter set to 5 epochs): this allowed to obtain good accuracy without incurring in overfitting. The batch size was fixed to 16. The model was fit with ADAM optimization algorithm and categorical cross-entropy as loss function. With respect to the model described in \cite{mymodel} some regularizers were added to avoid overfitting and make the network less sensitive to specific neurons weights. A L1-L2 kernel regularization was added to the two 1D convolutional layers $(l1=0.001, l2=0.002)$ and a L2 kernel regularizer $(l2=0.001)$ was added to the fully connected DNN layer; moreover, 0.5 dropouts were introduced.
\subsection{Long-Short-Term-Memory and Deep Neural Network}\label{LSTM}
The second model was implemented to test the possibility of generalizing over temporal sequences of variable length. To implement such an approach the data were padded with zeroes, as mentioned in the previous Section. Since the required masking layer was not supported by the Keras implementation of the CNN layer, we decided to opt for a simpler model: a 64 neurons LSTM, followed by a 32 neurons dense layer and a 2 neurons output layer with a sigmoidal activation function; also in this case L1-L2 regularization and 0.5\% dropout were used to avoid overfitting. The optimization algorithm, the loss function and the validation approach with early stopping were the same as before. 
By using this model the possibility of learning independently from the length of the temporal sequence was granted. This represents a further step towards the implementation of the same algorithm on the robot; indeed, no previous knowledge on the duration of the movement would be required to perform the classification, since the model is trained on variable temporal sequences.

\section{Results}\label{results}
Results for the classifiers performance are presented for both the weight and the carefulness features and for both the considered source of data: the motion capture and the optical flow from the robot camera.

\subsection{Carefulness level}  \label{carefSection} 
The performances in the classification of the carefulness level with the model presented in Sect. \ref{CNN} are reported in Table \ref{tab:carefRes}. 
\begin{table}[h]
\centering
\caption{Model accuracy (\%, mean and standard deviation) on carefulness level classification with the CNN-LSTM-DNN model. In brackets are the results when volunteer 8 was included in the data set\\}
\label{tab:carefRes}
\begin{tabular}{l|c|c}
         & \textbf{Motion capture}           & \textbf{Optical flow}\\
\hline
\hline
\textit{Training} & $92.15 (92.00)\pm2.14 (3.42)$ & $94.03 (92.18)\pm1.05 (1.00)$ \\
\textit{Test}     & $91.68 (90.97)\pm5.00 (11.12)$ & $90.54 (89.43)\pm6.56 (7.59)$ \\
\end{tabular}
\end{table}
\begin{figure}[H]
\begin{subfigure}{.5\textwidth}
  \centering
  \includegraphics[width=.99\linewidth]{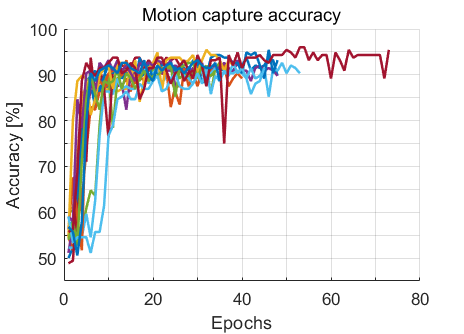}
  \caption{MoCap accuracy}
  \label{fig:moc_acc}
\end{subfigure}%
\begin{subfigure}{.5\textwidth}
  \centering
  \includegraphics[width=.99\linewidth]{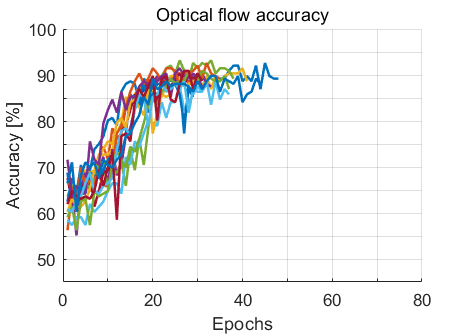}
  \caption{OF accuracy}
  \label{fig:of_acc}
\end{subfigure}
\caption{Accuracy in the carefulness classification with CNN-LSTM-DNN model for the validation set for each fold. Accuracy from motion capture (\ref{fig:moc_acc}) and from optical flow features (\ref{fig:of_acc})}
\label{fig:optoCaref}
\end{figure}
\noindent When performing the Leave-One-Out cross validation we noticed that the classification accuracy associated to volunteer 8 was significantly lower than the average (\textbf{MoCap test} \textit{all:} $90.97\pm11.12$ \textit{vol8:} 51.62; \textbf{OF test} \textit{all:} $89.43\pm7.59$ \textit{vol8:} 77.42). Examining the experiment videos we noticed that the volunteer 8 was particularly careful even when handling the glasses not containing water. Our impression was confirmed after computing the median duration of the not careful movements among the subjects. The duration for volunteer 8 ($2.04\pm0.18$ seconds, median and median absolute deviation) differed significantly from the ones of the other participants, as for the rest of the group the median duration resulted $1.47\pm0.15$ seconds (Kruskal-Wallis test: $\chi^{2}(14, N=480)=136.8, p<.01 $). In Table \ref{tab:carefRes} we have reported in brackets the results when including this subject in the dataset. As can be observed, when the participant was included the accuracy variance on the test increased significantly for each the sensing modalities.\\
Figure \ref{fig:optoCaref} shows the trend of the accuracy over the epochs for the validation set of each one of the folds. Comparing the graphs for the two sources of data ((a) motion capture, (b) optical flow) it can be noticed how the first one reaches an accuracy above the $80\%$ in less than 10 epochs, while, using the features from the optical flow, more training is necessary to reach the same level of accuracy (over 20 epochs). Furthermore, the accuracy trend of the motion capture features is more stable.\par
Similarly, the carefulness classification performance with the model presented in Sect. \ref{LSTM}, fed with the original temporal sequences of variable lengths, is shown in Table \ref{tab:carefRes_LSTM}. As before, the variability in the test accuracy reduced when volunteer 8 is excluded from the dataset, and the overall accuracy improved for both the sensing modalities. With this model, compared to the values in Table \ref{tab:carefRes}, the accuracy achieved with the MoCap data is higher, while the one of the OF slightly reduced.
\begin{table}[h]
\centering
\caption{Model accuracy (\%, mean and standard deviation) on carefulness level classification for simpler LSTM-DNN model. In brackets the results considering volunteer 8}
\label{tab:carefRes_LSTM}
\begin{tabular}{l|c|c}
         & \textbf{Motion capture}           & \textbf{Optical flow}\\
\hline
\hline
\textit{Training} & $96.57 (94.32)\pm1.19 (1.77)$ & $92.10 (90.39)\pm4.58 (2.56)$ \\
\textit{Test}     & $95.17 (92.66)\pm5.56 (8.49)$ & $88.38 (86.50)\pm8.68 (10.75)$ \\
\end{tabular}
\end{table}

\subsection{Weight} \label{Results_weight}
In Table \ref{tab:weiRes} are shown the results for the classification of the weight achieved with re-sampled data on the first implemented model. In this case, volunteer 8 did not present any peculiarity and therefore it was included in the dataset. As we can observe in Table \ref{tab:weiRes}, the accuracy with the motion capture data is above 60\% and is higher than the one obtained from the optical flow.
\begin{table}[ht]
\centering
\caption{Model accuracy (\%, mean and standard deviation) on weight classification with the CNN-LSTM-DNN model, fed with re-sampled data\\}
\label{tab:weiRes}
\begin{tabular}{l|c|c}
         & \textbf{Motion capture}           & \textbf{Optical flow}\\
\hline
\hline
\textit{Training} & $64.10\pm2.34$ & $55.24\pm2.37$ \\
\textit{Test}     & $61.83\pm7.16$ & $54.47\pm4.29$ \\
\end{tabular}
\end{table}
\\
\\Finally, Table \ref{tab:weightRes_LSTM} reports the accuracy for the weight classification with the LSTM-DNN model, fed with the original temporal sequences of different lengths. In this case the performance was comparable between the data from the two sensing modalities.
\begin{table}[h]
\centering
\caption{Model accuracy (\%, mean and standard deviation) on weight level classification for the second model, LSTM-DNN}
\label{tab:weightRes_LSTM}
\begin{tabular}{l|c|c}
         & \textbf{Motion capture}           & \textbf{Optical flow}\\
\hline
\hline
\textit{Training} & $54.95\pm2.66$ & $55.30\pm1.95$ \\
\textit{Test}     & $54.75\pm5.27$ & $53.29\pm3.59$ \\
\end{tabular}
\end{table}

\noindent We have noticed that, despite adopting the same approach, the accuracy on the weight classification is not as satisfying as the one achieved for the carefulness. A possible explanation of these results could be related to the different effect that weight may have on different transport movements. Possibly the weight influence varies if the transportation is from top to bottom or vice-versa. Furthermore, the presence of water in some of the glasses may have led the subjects to focus mainly on the carefulness feature, unconsciously overlooking the weight difference. Therefore, we add two specifications of the second hypothesis: \textit{\textbf{(H2.1)}} the influence of the weight during transportation is dependent on the trajectory of the motion; \textit{\textbf{(H2.2)}} when an object is associated with an high level of carefulness, the weight has a limited influence on the transportation movement. Both hypotheses were tested with the first model, which gave better results for the weight classification. Concerning the first hypothesis, we reduced the variability in the movements and tried to discriminate the weight in the subset of transport movements from the scale towards the shelves (\textbf{MoCap:} \textit{Tr}: $68.90\pm2.68$ \textit{Test}: $63.42\pm8.96$; \textbf{OF:} \textit{Tr}: $59.10\pm4.27$ \textit{Test}: $55.17\pm6.24$); there is a slight improvement for both the data sources compared to the values in Table \ref{tab:weiRes}. Notice that the trajectories still have a discrete amount of variability since the position to reach on the shelf could be left or right, high or low. The second hypothesis was investigated by testing the weight discrimination within the subset of objects which required the same carefulness level: low (\textbf{MoCap:} \textit{Tr}: $64.49\pm5.24$ \textit{Test}: $61.93\pm6.86$; \textbf{OF:} \textit{Tr}: $62.52\pm3.53$ \textit{Test}: $56.84\pm6.77$) or high (\textbf{MoCap:} \textit{Tr}: $62.72\pm3.65$ \textit{Test}: $59.03\pm8.73$; \textbf{OF:} \textit{Tr}: $57.92\pm1.31$ \textit{Test}: $53.48\pm7.63$). For both the tests the results are inconclusive, since the classification accuracies have not changed much respect to the ones reported in Table \ref{tab:weiRes}. It should be noted though that the dimension of the dataset used to validate hypotheses \textit{\textbf{(H2.1)}} and \textit{\textbf{(H2.2})} halved, which has an impact on the statistical relevance of the results.



\section{Discussion}
Regarding the carefulness feature, as reported in Table \ref{tab:carefRes} the first classifier is able to correctly discriminate if the transportation of the object requires carefulness or not, independently from the sensing modality used. Considering the performance on the data coming from the two sources, no significant difference is detected between them. Therefore, not only using an accurate system such as the motion capture, that integrates sensory inputs from different locations to estimate the position in space of the target, but also using the camera of the robot (single point of view), it is possible to extract features to discriminate between careful and not careful motions. Figure \ref{fig:optoCaref} shows an insight on how the learning process advanced for the two data sources. Even though the final performances are comparable, it can be appreciated how the model trained with the features from the motion capture converges quicker to an accuracy value above the 80\%.\\
The approach adopted with the second classifier is more general, in the sense that data are not re-sampled to have the same dimension but the variability in their duration is taken into account. Even though this model is simpler, with just one LSTM and one dense layer, the performance on the carefulness classification considering the MoCap data increased (see Table \ref{tab:carefRes_LSTM} for reference). Although the accuracy using the optical flow is slightly lower, we consider this as a promising step towards the implementation of the same algorithm on the robot. 
\par Concerning the weight, the accuracy achieved for both the sensing modalities and for both the models is lower than the one obtained for the carefulness (see Tables \ref{tab:weiRes} and \ref{tab:weightRes_LSTM} for reference). To explain this outcome in Sect. \ref{Results_weight} we have formalized two additional hypotheses. \textit{\textbf{(H2.1)}} was inspired by \cite{PalinkoSciutti}, where it has been proposed that the vertical component of the velocity during the manipulation of an object is perceived by humans as informative about its weight. Since the trials in our dataset explored a variety of directions and elevations, this introduced a great variability in the vertical component of the velocity. Instead, concerning \textit{\textbf{(H2.2)}}, we have supposed that the greatest challenge for the volunteers during the experiment is to safely handle the glasses full of water; the difference in weight between the objects was not remarkable in comparison with the stark contrast given by the presence (or absence) of water. As mentioned in Sect. \ref{Results_weight} the first classifier was tested against these hypotheses, but no significant improvements in the accuracy have been achieved. Given the results of our experiment we can not validate hypothesis \textit{\textbf{(H2)}}. However, since we have explored only a subset of the possible kinematic features we can not argue against this hypothesis either. A possibility for future works is to focus on the vertical component of the velocity. Furthermore, \textit{\textbf{(H2.1)}} and \textit{\textbf{(H2.2)}} should be explored on reasonably extended datasets to obtain more reliable results. 

\section{Conclusions}
As human-robot interactions are becoming increasingly frequent, it is crucial that robots gain certain abilities, such as the correct interpretation of implicit signals associated with the human movement. In this study we focused on two fundamental implicit signals commonly communicated in human movements: the impact of the weight and the carefulness required in the object manipulation (e.g. transport of fragile, heavy and unstable objects). Our hypotheses aimed to demonstrate that it is possible to discriminate between lighter and heavier items \textit{\textbf{(H2)}} and to infer the carefulness required by human operator in manipulating objects \textit{\textbf{(H1)}}. We proved that it is feasible to reliably discriminate when the human operator recruits motor commands of careful manipulation during the transportation of an object. Indeed, it is reliable to estimate extreme carefulness from two different typologies of sensory acquisition: from motion tracking system and from the single view point of the robot`s camera observing the movement. On the other hand, the proposed algorithms show lower accuracy when applied to weight classification, and these results does not allow us to validate our second hypothesis. The estimation of the weight from human motion should be subject of further studies, exploring other classification strategies or kinematic features subset (e.g. extraction of the vertical components of the velocity during manipulation).
This study firmly supports the research in human-robot interaction, especially in the direction of addressing complex situations in realistic settings (e.g.: industrial environment, construction site, home care assistance, etc.). In these specific scenarios the robot can autonomously leverage on insights inferred from implicit signals, such as the carefulness required to move a object, in order to facilitate the cooperation with the human partner. 
\section*{Acknowledgement}
This paper is supported by CHIST-ERA, (2014-2020) project InDex (Robot In-hand Dexterous manipulation).

%
%

\bibliography{references} 

\begin{thebibliography}{10}
\providecommand{\url}[1]{{#1}}
\providecommand{\urlprefix}{URL }
\expandafter\ifx\csname urlstyle\endcsname\relax
  \providecommand{\doi}[1]{DOI~\discretionary{}{}{}#1}\else
  \providecommand{\doi}{DOI~\discretionary{}{}{}\begingroup
  \urlstyle{rm}\Url}\fi

\bibitem{velWeight}
Bingham, G.: Kinematic form and scaling: further investigations on the visual
  perception of lifted weight.
\newblock Journal of experimental psychology. Human perception and performance
  \textbf{13}(2), 155—177 (1987)

\bibitem{keras}
Chollet, F., et~al.: Keras.
\newblock \url{https://keras.io} (2015)

\bibitem{legibility}
{Dragan}, A.D., {Lee}, K.C.T., {Srinivasa}, S.S.: Legibility and predictability
  of robot motion.
\newblock In: Proceedings of the 8\textsuperscript{th} ACM/IEEE International
  Conference on Human-Robot Interaction, pp. 301--308. Tokyo, Japan (2013)

\bibitem{farn:OF}
Farneb{\"a}ck, G.: Two-frame motion estimation based on polynomial expansion.
\newblock In: In: Proceedings of the 13\textsuperscript{th} Scandinavian
  Conference on Image Analysis, {LNCS} 2749, pp. 363--370. Gothenburg, Sweden
  (2003)

\bibitem{inpaintn}
Garcia, D.: Robust smoothing of gridded data in one and higher dimensions with
  missing values.
\newblock Computational Statistics and Data Analysis \textbf{54}, 1167--1178
  (2010)

\bibitem{yarp}
Metta, G., Fitzpatrick, P., Natale, L.: Yarp: Yet another robot platform.
\newblock International Journal of Advanced Robotic Systems \textbf{3} (2006)

\bibitem{lstm}
{Neverova}, N., {Wolf}, C., {Lacey}, G., {Fridman}, L., {Chandra}, D.,
  {Barbello}, B., {Taylor}, G.: Learning human identity from motion patterns.
\newblock IEEE Access \textbf{4}, 1810--1820 (2016)

\bibitem{PalinkoSciutti}
Palinko, O., Sciutti, A., Patane, L., Rea, F., Nori, F., Sandini, G.:
  Communicative lifting actions in human-humanoid interaction.
\newblock In: Proceedings of the 14\textsuperscript{th} IEEE-RAS International
  Conference on Humanoid Robots, pp. 1116--1121. Madrid, Spain (2015)

\bibitem{ActionObservation}
Press, C.: Action observation and robotic agents: Learning and
  anthropomorphism.
\newblock Neuroscience \& Biobehavioral Reviews \textbf{35}(6), 1410 -- 1418
  (2011)

\bibitem{Rea}
Rea, F., Vignolo, A., Sciutti, A., Noceti, N.: Human motion understanding for
  selecting action timing in collaborative human-robot interaction.
\newblock Frontiers in Robotics and AI \textbf{6}, 58 (2019)

\bibitem{mymodel}
{Sainath}, T.N., {Vinyals}, O., {Senior}, A., {Sak}, H.: Convolutional, long
  short-term memory, fully connected deep neural networks.
\newblock In: Proceedings of the 2015 IEEE ICASSP, pp. 4580--4584. Brisbane,
  Australia (2015)

\bibitem{sanchezGrip}
Sanchez, J., Corrales, J.A., Bouzgarrou, B.C., Mezouar, Y.: Robotic
  manipulation and sensing of deformable objects in domestic and industrial
  applications: a survey.
\newblock The International Journal of Robotics Research \textbf{37}(7),
  688--716 (2018)

\bibitem{billard:weight}
{Sanchez-Matilla}, R., {Chatzilygeroudis}, K., {Modas}, A., {Duarte}, N.F.,
  {Xompero}, A., {Frossard}, P., {Billard}, A., {Cavallaro}, A.: Benchmark for
  human-to-robot handovers of unseen containers with unknown filling.
\newblock IEEE Robotics and Automation Letters \textbf{5}(2), 1642--1649 (2020)

\bibitem{sandini}
Sandini, G., Sciutti, A.: Humane robots—from robots with a humanoid body to
  robots with an anthropomorphic mind.
\newblock ACM Transactions on Human-Robot Interaction \textbf{7}, 1--4 (2018)

\bibitem{sciutti:weight}
Sciutti, A., Patane, L., Nori, F., Sandini, G.: Understanding object weight
  from human and humanoid lifting actions.
\newblock Autonomous Mental Development, IEEE Transactions \textbf{6}, 80--92
  (2014)

\bibitem{grip}
{Su}, Z., {Hausman}, K., {Chebotar}, Y., {Molchanov}, A., {Loeb}, G.E.,
  {Sukhatme}, G.S., {Schaal}, S.: Force estimation and slip
  detection/classification for grip control using a biomimetic tactile sensor.
\newblock In: Proceedings of 15\textsuperscript{th} IEEE-RAS International
  Conference on Humanoid Robots, pp. 297--303. Seoul, Korea (2015)

\bibitem{vignolo:OF}
Vignolo, A., Noceti, N., Rea, F., Sciutti, A., Odone, F., Sandini, G.:
  Detecting biological motion for human–robot interaction: A link between
  perception and action.
\newblock Frontiers in Robotics and AI \textbf{4}, 14 (2017)

\bibitem{vignolo:OF2}
Vignolo, A., Rea, F., Noceti, N., Sciutti, A., Odone, F., Sandini, G.:
  Biological movement detector enhances the attentive skills of humanoid robot
  icub.
\newblock In: Proceedings of the 16\textsuperscript{th} IEEE-RAS International
  Conference on Humanoid Robots. Cancun, Mexico (2016)

\end{thebibliography}
\bibliographystyle{spmpsci}

\end{document}